\documentclass[conference]{IEEEtran}
\IEEEoverridecommandlockouts

\usepackage{graphicx}
\usepackage{subfigure}
\usepackage{stfloats}
\usepackage{textcomp}
\usepackage{balance}
\usepackage{amsmath}
\usepackage{amsfonts,amssymb}
\usepackage{bm}
\usepackage{extarrows}
\usepackage[colorlinks=true,
            linkcolor=blue,
            anchorcolor=black,
            citecolor=black,
            ]{hyperref}
\usepackage[margin=1.9cm]{geometry}
\usepackage[linesnumbered,ruled,vlined]{algorithm2e}
\usepackage{cite}
\usepackage{url}
\usepackage{multirow}  
\usepackage{array}    
\title{\LARGE \bf
Capsizing-Guided Trajectory Optimization for Autonomous Navigation with Rough Terrain
}

\author{Wei Zhang$^{1,\dag}$, Yinchuan Wang$^{1,\dag}$, Wangtao Lu$^{2}$, Pengyu Zhang$^{1}$, Xiang Zhang$^{1}$, Yue Wang$^{2}$, Chaoqun Wang$^{1,*}$
\thanks{ This work was supported by Key R\&D Program of Shandong Province under Grant No. 2024CXGC010210, and in part by TaiShan Youth Scholar Scheme of Shandong Province, China.}
\thanks{$^{1}$Wei Zhang, Yinchuan Wang, Pengyu Zhang, Xiang Zhang, Chaoqun Wang are with the School of Control Science and Engineering, Shandong University, \#17923, Jingshi Road, Jinan, Shandong Province, China. Chaoqun Wang is the corresponding author. Email: \tt \{David\_Zhang, sdwyc, sucro\_zhangxiang, pengyu3z3\}@mail.sdu.edu.cn, chaoqunwang@sdu.edu.cn}
\thanks{$^{2}$Wangtao Lu, Yue Wang are with the State Key Laboratory of Industrial Control Technology and Institute of Cyber-Systems and Control, Zhejiang University, Hangzhou, China. Email: \tt 16622808052@163.com, ywang24@zju.edu.cn}
\thanks{$^{\dag}$The first two authors contributed equally to this work.}%
}

\begin{document}
\newgeometry{left=1.9cm, right=1.9cm, top=2.54cm, bottom=1.9cm}
\maketitle
\thispagestyle{empty}
\pagestyle{empty}
\begin{abstract}
It is a challenging task for ground robots to autonomously navigate in harsh environments due to the presence of non-trivial obstacles and uneven terrain. This requires trajectory planning that balances safety and efficiency. The primary challenge is to generate a feasible trajectory that prevents robot from tip-over while ensuring effective navigation. In this paper, we propose a capsizing-aware trajectory planner (CAP) to achieve trajectory planning on the uneven terrain. The tip-over stability of the robot on rough terrain is analyzed. Based on the tip-over stability, we define the traversable orientation, which indicates the safe range of robot orientations. This orientation is then incorporated into a capsizing-safety constraint for trajectory optimization. We employ a graph-based solver to compute a robust and feasible trajectory while adhering to the capsizing-safety constraint. Extensive simulation and real-world experiments validate the effectiveness and robustness of the proposed method. The results demonstrate that CAP outperforms existing state-of-the-art approaches, providing enhanced navigation performance on uneven terrains.

\end{abstract}

\section{Introduction}
The rapid advancement of mobile robotics has enabled the development of robots capable of navigating and performing tasks in increasingly challenging environments, such as planet exploration \cite{chang2022lamp}, disaster relief \cite{hariprasath2024path}, etc. In these scenarios, robots are required to operate autonomously in harsh, unpredictable, and uneven terrains. Specifically, there are many non-strict obstacles that affect the stability and safety of the robot, such as slopes and potholes. It is challenging for wheeled robots to maintain stability and safety when navigating through rough environments. 

Uneven surfaces pose a risk to the stability of the robot, with concerns about tip-overs, which can lead to navigation failure and damage to the hardware system. These challenges arise from the need for the robot to maintain stability and safety while adapting to changes in the terrain. In response to these challenges, researchers propose various approaches to address the stable and efficient motion planning of wheeled robots in rough terrain. However, existing solutions often focus on planning a trajectory that emphasizes either efficiency or safety, but rarely both, especially in rugged environments that can severely impact robot safety and stability. 

\begin{figure}
    \centering
    \includegraphics[width=1\linewidth]{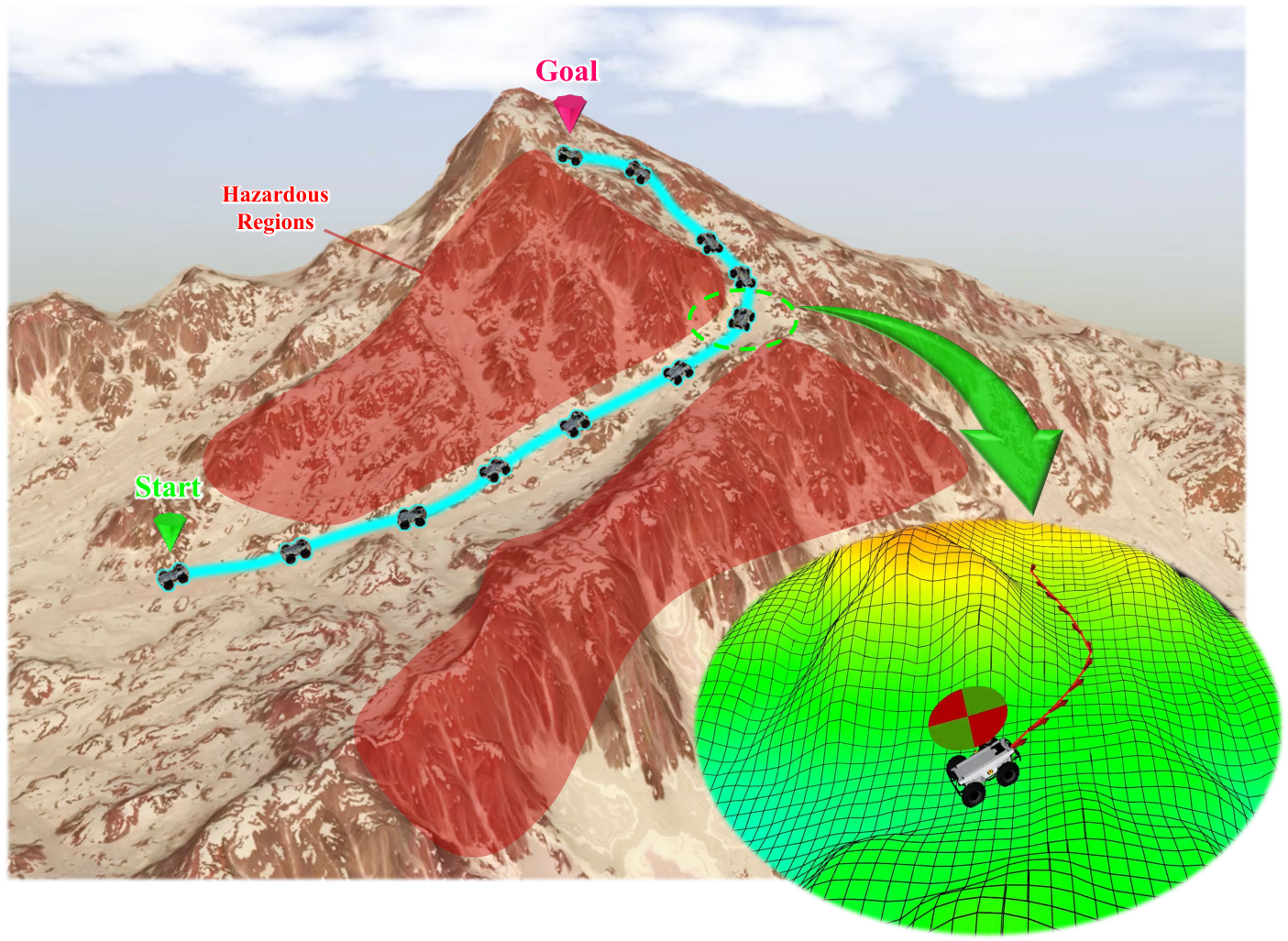}
    \vspace{-0.8cm}
    \caption{A ground robot navigates on the rough terrain with the capsizing-aware trajectory planner. The blue line is the navigation trajectory. The right bottom figure shows a snapshot of the navigation, which exhibits the generated trajectory (red line) and poses of the robot (red arrows). The colored pie chart reflects the traversable orientation of the robot.}
    \label{fig: title_figure}
\end{figure}

Several methods \cite{jian2022putn,wang2024hap} expand the RRT algorithm and adapt it to the uneven terrain. Their methods construct a tree structure on the ground and rapidly seek a safe trajectory from the tree, but their trajectories are coarse and infeasible. Xu \textit{et al.} \cite{xu2023uneven_planner} utilize the 2D-to-3D state space mapping and propose a trajectory optimization method to facilitate the trade-off between safety and efficiency. Some studies explore the terrain-aware-based method, aiming to offer a perception of terrain for trajectory planning, such as \cite{atas2022elevation, cai2023probabilistic}. Leininger \textit{et al.} \cite{abe2024gpr} exploits the Gaussian process regression (GPR) to generate a traversability map and then uses the RRT* algorithm to achieve the trajectory planning. Wang \textit{et al.} \cite{wang2023towards} formulate a penalty field to constrain the robot moving on the flat terrain, which is further introduced to a trajectory optimization for obtaining a safe trajectory. However, these methods rely on guidance from perceived maps, they are challenging to achieve trajectory planning when the perceived maps are inaccurate.

With the aid of deep learning technique, Shen \textit{et al.} \cite{shen2023efficient} exploits a neural network to train a safe cost function of the edge and build topology-aware RRT to generate a feasible local trajectory. \cite{yu2024real,xiao2021learning,siva2024self} focus on achieving end-to-end trajectory planning by deep reinforcement learning (DRL). Nevertheless, it remains challenging for these learning-based methods to effectively adapt to uneven terrains in a variety of real-world scenarios.


In this paper, we present a trajectory planner, \textbf{C}apsizing-\textbf{A}ware \textbf{P}lanner (CAP), for wheeled robots navigating uneven terrain, with a specific focus on preventing tip-overs while optimizing the robot’s motion as shown in Fig. \ref{fig: title_figure}. We leverage the stability pyramid model to derive the traversability of orientations when the robot moves on the ground. Subsequently, we formulate a trajectory optimization paradigm, incorporating a capsizing-safety constraint established from the traversable orientation. Furthermore, we introduce a robust solver based on graph optimization that balances multiple constraints, resulting in a trajectory that is both stable and efficient. We validate our approach through a series of experiments, the results demonstrate its effectiveness and high performance in both simulated and real-world rough terrain scenarios. The video demonstration of conducted experiments is available at link\footnote{Video demonstration: \href{https://youtu.be/81joZKvpmck}{youtu.be/81joZKvpmck} }. In summary, the contributions of the proposed CAP are as follows:
\begin{itemize}
    \item We analyze the tip-over stability of the robot under uneven terrain. The traversable orientation is then derived by the tip-over stability, which reflects the condition of state transition when the robot is on the ground. 

    \item Based on the traversable orientation, a capsizing-safety constraint is established for trajectory optimization of the ground robot, which is further integrated into the object function of the trajectory optimization.

    \item We design a robust solver for trajectory optimization with the established capsizing-safety constraint. The solver exploits the graph optimization model to trade off constraints and obtains a smooth and safe trajectory.


    
    
    
\end{itemize}
The remainder of this article is structured as follows: 
The details of the proposed method are introduced in Sec. \ref{sec: methodology}. Subsequently, the conducted experimental evaluation is presented in Sec. \ref{sec: experiments}. We draw conclusions and discussion of future work in Sec \ref{sec: conclusion}.

\begin{algorithm}[tp]
\caption{Capsizing-Aware Planner}
\label{alg: pipeline}
\LinesNumbered
\KwIn{2.5D grid map $\mathcal{M}_t$, Global Path $\sigma_t$\;}
\KwOut{Trajectory $\mathcal{S}_t$, Control Signal $\mathbf{u}_t$\;}

$\mathbf{g}^{*} \leftarrow \mathbf{x}^{\sigma}_n$;  $\rhd$ Initialization \\
$\mathcal{S}_t \leftarrow \emptyset,\ \mathbf{u}_t \leftarrow \mathbf{0}$\;

\While{$\left\| \mathbf{x}_r - \mathbf{g}^{*} \right\| > \eta$}{
    \eIf{$\nexists \mathcal{S}_{t-1}$}
    {$\mathcal{S}_t \leftarrow \mathbf{LineToGoal}(\mathbf{x}_r, \mathbf{g}^{*})$\;}
    {$\mathcal{S}_t \leftarrow \mathcal{S}_{t-1}$\;}
    $\mathcal{F}_{GN} \leftarrow \mathbf{SurfaceRegression}(\mathcal{M}_t)$\;
    \For{$k \in N_{iter}$}{
        $\mathcal{C}_t \leftarrow \mathbf{ConstraintBuild}(\mathbf{x}_i, \mathcal{F}_{GN}, \mathcal{S}_t)$\;
        $\mathtt{solver} \leftarrow G(\mathcal{C}_t, \mathcal{S}_t)$\;
        $\delta S \leftarrow \mathtt{solver}.\mathbf{solve}(G(\mathcal{C}_g, \mathcal{S}_t))$\;
        $\mathcal{S}_t \leftarrow \mathcal{S}_t + \delta S$\;
    }
    $\mathbf{u}_t \leftarrow \mathbf{TrackControl}(\mathbf{x}_r, \mathcal{S}_t)$\;
}
\end{algorithm}

\section{Methodology}
\label{sec: methodology}
\subsection{Pipeline}
Generally, a ground robot on the uneven terrain works in the 2.5D space $\mathbb{W}$, whose state $\mathbf{x} \in \mathbb{W}$ can be represented as $\mathbf{x} = [x,y,z,\theta]$. Here, $[x,y,z]$ is the position of the robot and $\theta$ represents the orientation. We denote $\mathcal{M}$ as the local grid map while the grid $m_i \in \mathcal{M}$ stores the ground height of terrain. Based on the grid map, a motion planning task is divided into two consecutive modules: trajectory planning and trajectory tracking. The trajectory planning module refines the global path $\sigma = \{\mathbf{x}^{\sigma}_{0}, \mathbf{x}^{\sigma}_{1}, \dots, \mathbf{x}^{\sigma}_{n}\}$ to generate the smooth and feasible trajectory $\mathcal{S}$. The generated trajectory is then tracked by the trajectory tracking module. 

The pipeline of the CAP is described in Alg. \ref{alg: pipeline}. At a navigation stance $t$, the CAP receives the 2.5D grid map $\mathcal{M}_t$ and global path $\sigma_t$ from the local mapper \cite{fankhauser2018elevation} and global path planner, respectively. The replanning procedure is constantly executed until the robot arrives at the goal $\mathbf{g}^{*}$ as shown in Lines 3-13 of Alg. \ref{alg: pipeline}. $\eta$ is the tolerance of navigation accomplishment. 

For one replanning iteration, the trajectory $\mathcal{S}_t$ is initialized by the last trajectory $\mathcal{S}_{t-1}$. if $\mathcal{S}_{t-1}$ is not exist, each state $\mathbf{x}^{\mathcal{S}}_i \in \mathcal{S}_t$ is determined by linear interpolation along the direction from $\mathbf{x}_r$ to $\mathbf{g}^{*}$ if $\mathcal{S}_{t-1}$, which is indicated by $\mathbf{LineToGoal}(\cdot)$ in Line 5 of Alg. \ref{alg: pipeline}. Subsequently, we use the interpolation method to fit the ground surface in $\mathcal{M}_t$, and obtain the field of ground normal $\mathcal{F}_{GN}$. A trajectory optimization process is then utilized to obtain the smoothest and safest trajectory in Lines 9-13 of Alg. \ref{alg: pipeline}. Here, we construct a graph-based $\mathtt{solver}$ to solve the optimization problem. The $\mathtt{solver}$ can calculate the updates $\delta S$ in terms of $\mathcal{C}_t$ and $\mathcal{S}_t$ as shown in Line 12 of Alg. \ref{alg: pipeline}. Based on the optimized trajectory $\mathcal{S}_t$, we adopt the nonlinear MPC strategy, detailed in \cite{kunhe2005lmpc}, to track the certain length of trajectory segment $S \in \mathcal{S}_t$. $\mathbf{u}_t$ is the received control signal of the robot in Line 14 of Alg. \ref{alg: pipeline}.
 

\subsection{Stability Analysis and Traversable Orientation}



\begin{figure}[t]
    \centering
    \includegraphics[width=1.0\linewidth]{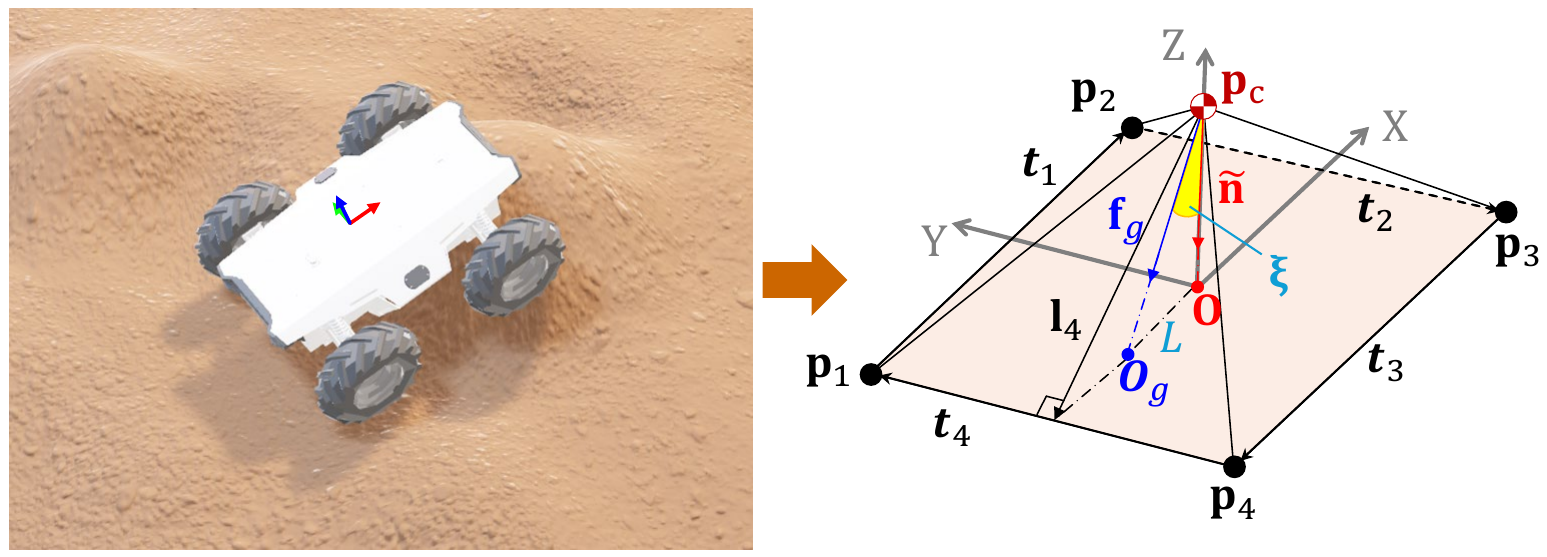}
    \caption{Stability pyramid model of the robot in uneven terrain. $\mathbf{p}_c$ represents the mass center of the robot. The orange regions represent the stability polygon surrounded by the contact points $\mathbf{p}_i, i\in\{1,2,3,4\}$. $\mathbf{f}_g$ and $\tilde{\mathbf{n}}$ are the gravity vector and reverse of the normal vector of $\mathcal{R}$, respectively.}
    \label{fig: stability pyramid}
\end{figure}

Given a wheeled robot on uneven terrain, it is assumed that the wheels still contact the ground surface and do not experience slip. We use the stability pyramid model \cite{wang2020stable} to judge whether the robot happens to capsize. The robot can be modeled as a $N$-gonal pyramid where $N$ is the number of contact points $\mathbf{p}_{i}(i\in \{1,\dots, N\})$ as shown in Fig. \ref{fig: stability pyramid}. We denote the $\mathcal{R}$ as the stability polygon surrounded by contact points $\mathbf{p}_i$, which is shown as the orange region in Fig. \ref{fig: stability pyramid}. The $\mathbf{t}_i$ represents the edge of $\mathcal{R}$, it can be calculated by $\mathbf{t}_i = {\mathbf{p}}_{i+1} - {\mathbf{p}}_{i} , i \in \{1, 2, 3\}$. Specifically ${\bf{t}}_4 = {\bf{p}}_{1} - {\bf{p}}_{4}$. $\mathbf{p}_c$ is the mass center of the robot. $\mathbf{f}_g$ and $\mathbf{n}$ represent the gravity vector and the normal vector of $\mathcal{R}$, respectively, i.e. $\tilde{\mathbf{n}} = -\mathbf{n}$. $\mathbf{O}$ and $\mathbf{O}_g$ are the projections of $\mathbf{f}_g$ and $\tilde{\mathbf{n}}$ onto the $\mathcal{R}$.
On uneven terrain, the robot maintains self-stability and avoids tipping over when the following conditions are satisfied:
\begin{equation}
\label{eq: stability_pyramid}    
\begin{aligned}
    &\min(\epsilon_i \arccos (\hat{\mathbf{f}}_g \cdot \hat{\mathbf{l}}_i)) > 0,\ i = \{1,2,3,4\}, \\[1mm]
&\begin{aligned}
     \mathit{s.t.} ,\ \mathbf{l} _i &= (\mathbf{I} -\hat{\mathbf{t}}_i\hat{\mathbf{t}}_{i}^{\top} )(\mathbf{p} _{i+1}-\mathbf{p}_c),\\
     \epsilon_i &= \left\{\begin{matrix}
    \begin{aligned}
     &+1,\ (\hat{\mathbf{l}}_i \times \hat{\mathbf{f}}_g) \cdot \hat{\mathbf{t}}_i < 0\\
     &-1,\ \mathrm{otherwise}
    \end{aligned}
    \end{matrix}\right.
\end{aligned}
\end{aligned}
\end{equation}
where $\mathbf{l}_i$ denotes the tip-over normal axis, which intersects with $\mathbf{p}_c$ and perpendicular to $\mathbf{t}_{i}$. The top script `` $\hat{ }$ " represents the normalization of a vector. $\mathbf{I}$ is the unit matrix. Eq. \ref{eq: stability_pyramid} indicates the robot can maintain stability and avoid tipping over when $\mathbf{O}_g$ lies within $\mathcal{R}$, i.e., $\mathbf{O}_g \in \mathcal{R}$. 

\begin{figure}[t]
        \centering
        \includegraphics[width=1.0\linewidth]{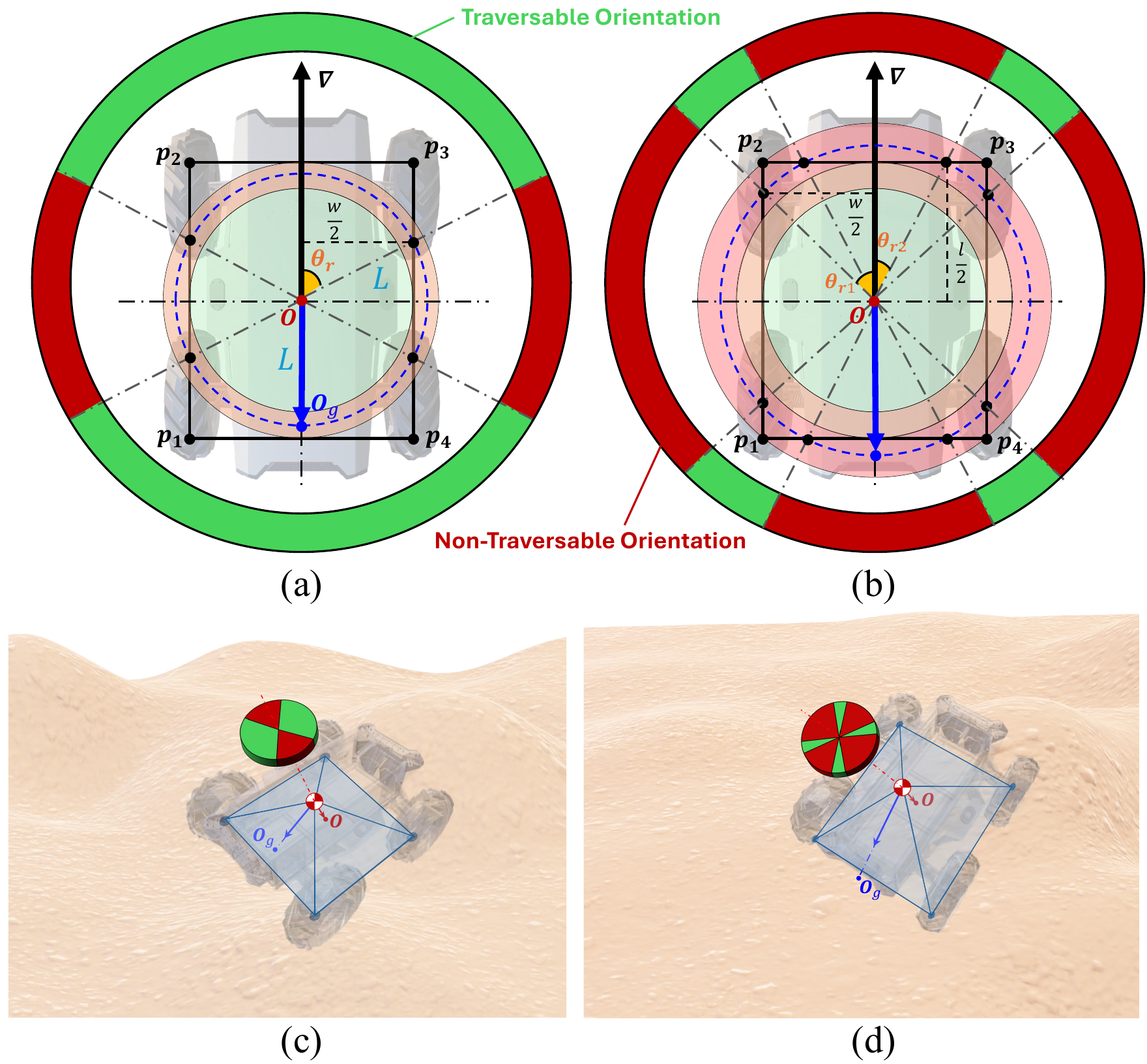}
        \caption{Traversable orientations of the robot on the uneven terrain. (c) and (d) shows two scenes of the robot on the ground. The blue pyramid is the stability pyramid of the robot. The color pies indicate the traversability of orientations. Here the reds and greens represent the traversable and non-traversable orientations, respectively. The traversability of orientations are derived in (a) and (b).}
        \label{fig: traversable_orientation}
\end{figure}

Based on the above model, at the same location of uneven terrain, the robot has different terrain traversability regarding different orientations. For the sake of brevity, we parameterized the wheeled robot as $[w, l, h]$ representing the width, length, and height of the mass center, respectively. The projection $\mathbf{O}$ of mass center $\mathbf{p}_c$ is located at the geometric center of the polygon $\mathcal{R}$. In Fig. \ref{fig: stability pyramid}, the projection length $L$ of $\overline{\mathbf{P}_c \mathbf{O}_g}$ can be written as
\begin{equation}
    L = h \tan \xi = h \tan \left(\arccos (\hat{\mathbf{f}}_g - \hat{\tilde{\mathbf{n}}}) \right),
\end{equation}
where $\xi$ is the stability angle. The traversability of orientation is related to the projection length $L$ as shown as blue arrows in Fig. \ref{fig: traversable_orientation}. When $L < \frac{w}{2}$, the projection point $\mathbf{O}_g$ of $\mathbf{f}_g$ is permanently lied inside on the green circle in Fig. \ref{fig: traversable_orientation}(a). This indicates that the robot can steer to any orientation at this location. However, as $L$ increases, the range of traversable orientation becomes increasingly restricted. In Fig. \ref{fig: traversable_orientation}(a), $\mathbf{O}_g$ is lied in the orange region when $\frac{w}{2} \le L < \frac{l}{2}$. $\nabla$ represents the gradient pointing in the direction of maximum increase of the $\mathcal{R}$. The traversability of orientation is shown as the colored ring whose red and green parts represent traversable and non-traversable orientations, respectively. According to the triangle relationship, the traversable orientation can be derived from the central angle $\theta_r$, which is calculated by
\begin{equation}
    \theta_r = \arcsin \frac{w}{2L},
\end{equation}
Thus, the traversable orientation is following 
\begin{equation}
 (\left \| \theta_{\scriptscriptstyle \nabla} \right \| < \theta_r ) \vee
(\pi - \theta_r < \left \| \theta_{\scriptscriptstyle \nabla} \right \| < \pi)
\end{equation}

Here, $\theta_{\scriptscriptstyle \nabla}$ is the robot orientation on the $\mathcal{R}$, described in polar coordinates relative to the $\nabla$. In Fig. \ref{fig: traversable_orientation}(b), $\mathbf{O}_g$ is lied in the oranges region when $\frac{l}{2} \le L < \frac{\sqrt{w^2+l^2}}{2}$. Following the similar routine with Fig. \ref{fig: traversable_orientation}(a), The traversable orientation can be derived from $\theta_{r1}$ and $\theta_{r2}$ in Fig. \ref{fig: traversable_orientation}(b). We have
\begin{equation}
\begin{aligned}
\theta_{r1} & = \arcsin \frac{w}{2L},\\
\theta_{r2} & = \arccos \frac{l}{2L}.
\end{aligned}
\end{equation}
The traversable orientation can be described as
\begin{equation}
 (\theta_{r2} < \left \| \theta_{\scriptscriptstyle \nabla} \right \| < \theta_{r1}) \vee
(\pi - \theta_{r1} < \left \| \theta_{\scriptscriptstyle \nabla} \right \| < \pi - \theta_{r2})
\end{equation}
If $L \ge \frac{\sqrt{w^2+l^2}}{2}$, $\mathbf{O}_g$ is outside of $\mathcal{R}$, which indicates the robot cannot traverse there whatever orienting to any direction. Overall, we summarize the traversable orientation can be determined by $H(\theta_{\scriptscriptstyle \nabla})$, which is equal to
{\footnotesize
 
\begin{equation}
\label{eq: traversable_orientation}
\left\{\begin{matrix}
\begin{aligned}
&h[\pi - \| \theta_{\scriptscriptstyle \nabla} \|]  &&, L <\frac{w}{2} \\
&\begin{aligned}
&h[\theta _r - \| \theta_{\scriptscriptstyle \nabla} \|] + \\[1mm]
&h[\| \theta_{\scriptscriptstyle \nabla} \| - (\pi - \theta _r)]\cdot h[\pi - \| \theta_{\scriptscriptstyle \nabla} \|] \end{aligned}&&, \frac{w}{2} \le L<\frac{l}{2} 
\\[1.5mm]
& \begin{aligned}
 &h[\theta_{r1} - \| \theta_{\scriptscriptstyle \nabla} \|]\cdot h[\| \theta_{\scriptscriptstyle \nabla} \| - \theta _{r2}] + \\[1mm]
 &h[\| \theta_{\scriptscriptstyle \nabla} \| - (\pi - \theta_{r1})] \cdot h[(\pi - \theta _{r2}) - \| \theta_{\scriptscriptstyle \nabla} \|]
\end{aligned} &&, \frac{l}{2} \le L<\frac{\sqrt{w^2+l^2}}{2} 
 \\& 0 &&, L \ge \frac{\sqrt{w^2+l^2}}{2}
\end{aligned}
\end{matrix}\right.
\end{equation}
}
$h(\cdot)$ represents the Heaviside function. When $H(\theta_{\scriptscriptstyle \nabla})>0$, the robot orientation belongs to traversable orientation. Eq. \ref{eq: traversable_orientation} reflects the constraints of the robot orientation $\theta$ on the 2D state space. These constraints are further introduced to the trajectory optimization procedure.

\subsection{Capsizing-Safety Constraint and Trajectory Optimization}
A wheeled robot is constrained on the ground surface, its pose $\mathbf{X} \in SE(3)$ can be uniquely represented by the state $\mathbf{x} \in \mathbb{W}$. We define $\mathfrak{g}(x,y)$ as the mapping of the grid map $\mathcal{M}$, which is mapping from the 2D position $[x,y] \in \mathbb{R}^{2}$ to the elevation $z \in \mathbb{R}$. Thus, given a robot pose $\mathbf{s}=[\mathbf{R}\ |\ \mathbf{p}]$, where $\mathbf{p}=[x,y] \in \mathbb{R}^{2}$ and $\mathbf{R}=[\mathbf{b}_x, \mathbf{b}_y] \in SO(2)$, we have 
\begin{equation}
\label{eq: se2_mapping}
\begin{aligned}
& z=\mathfrak{g}(x,y); &&\mathbf{b}_z = \mathbf{n} ;\\
& \mathbf{b}_x =  \frac{(\mathbf{e}_z \times \mathbf{b}_{yaw}) \times \mathbf{b}_z}{\left \| (\mathbf{e}_z \times \mathbf{b}_{yaw}) \times \mathbf{b}_z \right \|}; &&\mathbf{b}_y = \mathbf{b}_z \times \mathbf{b}_{x}.
\end{aligned}
\end{equation}
$\mathbf{b}_x, \mathbf{b}_y, \mathbf{b}_z$ are the unit vectors aligned with the three coordinate axes of the robot's base. $\mathbf{n}$ is the unit normal vector of the stability polygon $\mathcal{R}$, which can be derived from the position of contact points using the SVD fitting method \cite{tufts1982singular}. $\mathbf{e}_z$ is the unit vector aligned with the \textit{z}-axis of global frame. $\mathbf{b}_{yaw}$ is the projection vector of $\mathbf{b}_x$ on the $XY$ plane, i.e., $\mathbf{b}_{yaw} = [\sin \theta, \cos \theta, 0]$. Eq. \ref{eq: se2_mapping} establish a state mapping $\mathfrak{f}: \mathbf{s}\in SE(2) \mapsto \mathbf{s}^{+} \in SE(3)$, where the 3D state $\mathbf{s}^{+} = [\mathbf{R}^{+} | \mathbf{p}^{+}]$, $\mathbf{R}^{+}=[\mathbf{b}_x, \mathbf{b}_y, \mathbf{b}_z]$, $\mathbf{p}^{+} = [x,y,z]^{\top}$. This indicates that we can achieve the 3D trajectory planning upon the ground surface by utilizing the 2D state $\mathbf{s} \in SE(2)$.
\begin{figure}
        \centering
        \includegraphics[width=0.95\linewidth]{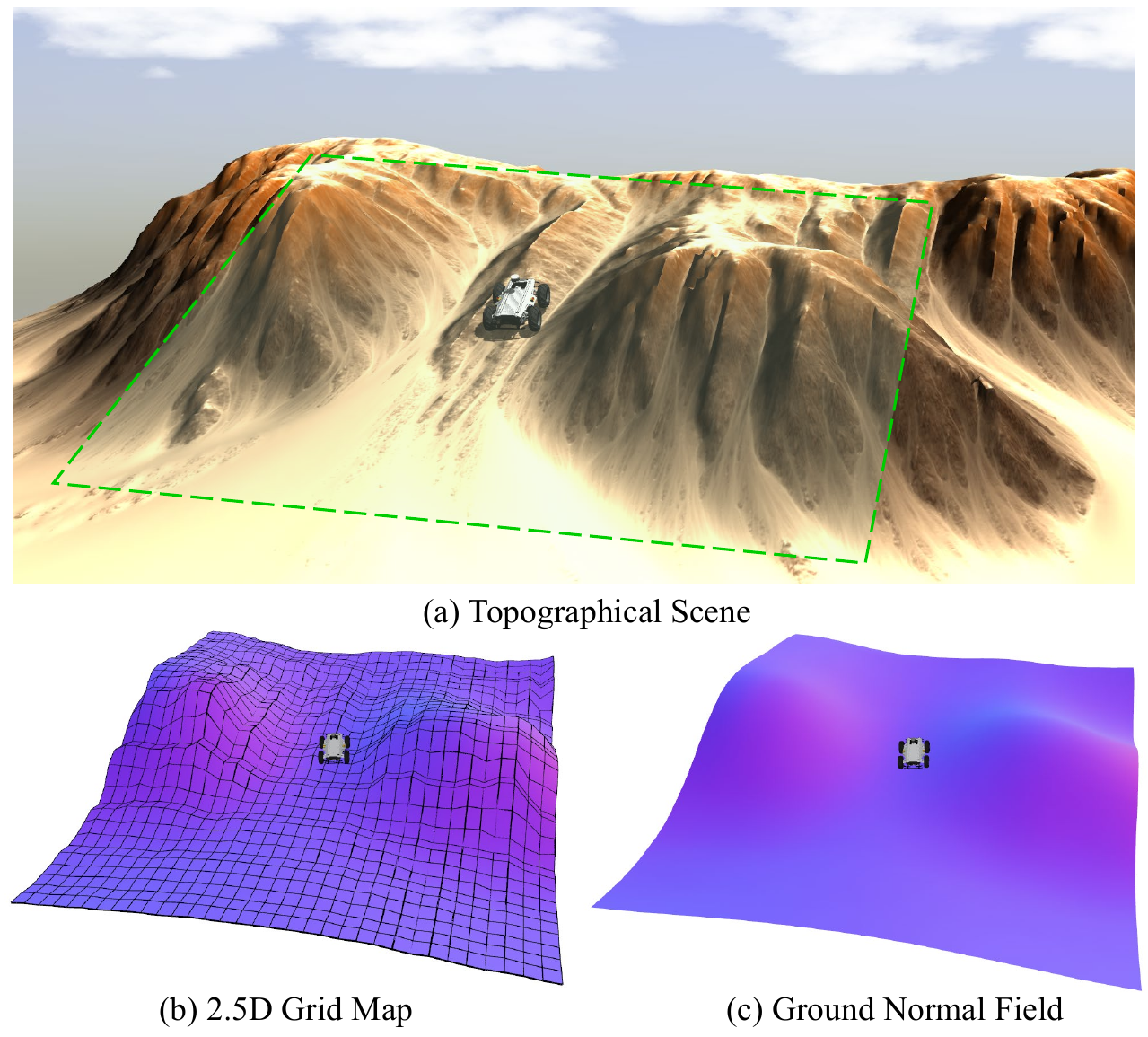}
        \caption{Ground normal field in an uneven terrain. (a) shows a topographical scene of a robot on uneven terrain. The green square represents the range of the grid map $\mathcal{M}$. (b) and (c) showcase the 2.5D grid map $\mathcal{M}$ and the ground normal field $\mathcal{F}_{GN}$, respectively. The colors of (b) and (c) reflect the normal direction.}
        \label{fig: gnf}
\end{figure}

To represent continuous 3D state $\mathbf{s}^{+}$ inside a $m \in \mathcal{M}$, we formulate a ground normal field $\mathcal{F}_{GN}$ to obtain the continuous representation of elevation and normal vector. Since directly adopting the regression algorithm is computationally expensive, we adopt a local regression window $\mathcal{B}$ comprising the grid $m \in \mathcal{M}$ and its eight-neighborhood to obtain $\mathcal{F}_{GN}$. Given the elevation $z$ and normal vector $\mathbf{n}=[n_x,n_y,n_z]$ stored in the $m$, we denote $\mathbf{v} = [z, n_x,n_y,n_z]$ as an embedding vector of a grid $m$ and $\mathcal{F}_{GN}(x,y) = \mathfrak{f}(\{\mathbf{v}_i\})$, $\mathbf{v}_i \in \mathcal{B}$. For a location $\mathbf{p}=[x,y] \in m$, we use the linear interpolation to nine estimation vectors $\{\tilde{\mathbf{v}}_i\}^{8}_{i=0}$ of $\mathbf{p}$ from $\{\mathbf{v}_i\}^{8}_{i=0}$. The each of $\tilde{\mathbf{v}}_i$ is calculated by
\begin{equation}
    \tilde{\mathbf{v}}_i = \frac{\|\mathbf{p} - \mathbf{p}_c\|}{\|\mathbf{p}^{i}_c - \mathbf{p}_c\|} \mathbf{v}_i,\ i \in \{0, 1, \dots, 8\},
\end{equation}
where $\mathbf{p}^{i}_c$ is the central coordinate of $m_i \in \mathcal{B}$, specially $\mathbf{p}_c$ is the central coordinate of $m_i$. Then, the ordinary Kriging algorithm \cite{oliver1990kriging} is exploited to predict $\mathbf{v}$ of the location $\mathbf{p}$, which can be estimated by
\begin{equation}
\label{eq: kriging_pre}
\begin{aligned}
    &\hat{\mathbf{v}} = \sum_{i=0}^{8} diag[\bm{\lambda}] \cdot \tilde{\mathbf{v}}_i = \sum_{i=0}^{8} \Lambda_i \tilde{\mathbf{v}}_i,\\
    &\mathit{s.t.,}\ \sum_{i=0}^{8} \Lambda_i=\mathbf{I}.
\end{aligned}
\end{equation}
$\bm{\lambda} = [\lambda_z,\lambda_{nx},\lambda_{ny},\lambda_{nz}]$ is the Kriging weight vector. $diag[\cdot]$ is a diagonal matrix with $\star$ as its diagonal element. $\lambda_z$, $\lambda_{nx}$, $\lambda_{ny}$, $\lambda_{nz}$ are the weight of $\mathbf{v}$. $\mathbf{I}$ is the identity matrix. Based on Kriging algorithm, $\hat{\mathbf{v}}$ is represented by the sum of the mean vector $\bar{\mathbf{v}}$ and bias $R$. We define the mean vector $\bar{\mathbf{v}}$ and the covariance $\bm{\Omega}_R$ as
\begin{equation}
    \bar{\mathbf{v}} = \frac{\sum_{i=0}^{8} \tilde{\mathbf{v}}_i}{9},\ \bm{\Omega}_R = \left [\sum_{i=0}^{8} \left ( \tilde{\mathbf{v}}_i-\bar{\mathbf{v}} \right )\right ]^{\otimes } ,
\end{equation}
where $[\star]^{\otimes}$ represents the tensor product of $\star$. The covariance $\mathbf{J}$ of the error $[\bar{\mathbf{v}}-\mathbf{v}]$ can be written as
\begin{align}
    \mathbf{J} 
    \label{eq: kriging_2}
    &= \sum_{i=0}^{8} \Lambda_i \mathbf{\Omega}_i \Lambda^{\top}_i+ \bm{\Omega}_{\mathbf{v}} - 2\sum_{i=0}^{8} \Lambda_i \mathbf{\Omega}_{i \mathbf{v}}\\
    \label{eq: kriging_3}
    &\xlongequal[]{\mathbf{R}_{\star}=\bm{\Omega}_\mathbf{R}-\bm{\Omega}_{\star}} 2 \sum_{i=0}^{8} \Lambda_i \mathbf{R}_{i \mathbf{v}} - \sum_{i=0}^{8} \Lambda_i \mathbf{\Omega}_i \Lambda^{\top}_i - \mathbf{R}_{\mathbf{v}}.
\end{align}
For simplicity, $\bm{\Omega}_{\star}$ represents the covariance of the vector $\star$, $\mathbf{\Omega}_{i \mathbf{v}}$ is the covariance $Cov(\mathbf{v}_i, \mathbf{v})$. Eq. \ref{eq: kriging_2} is obtained by combining with Eq: \ref{eq: kriging_pre}. The top equation of the ``$=$'' is the semivariogram in the Kriging algorithm, which is combined to obtain Eq. \ref{eq: kriging_3}. Thus, we establish an optimization function to get the optimal Kriging weight set $\{\bm{\lambda}_i\}^{8}_{i=0}$, which can be written as
\begin{equation}
\label{eq: kriging_opt}
\begin{aligned}
    &\bm{\lambda}^{*} = \mathop{\arg\min} \left ( \mathbf{J} + 2 \bm{\Phi} \left (\sum_{i=0}^{8} \Lambda_i-\mathbf{I} \right )   \right ),
\end{aligned}
\end{equation}
where $\bm{\Phi}$ is a matrix of Lagrange multiplier. To solve the Eq. \ref{eq: kriging_opt}, we set the Jacobian matrix of the objective function equal to the zero matrix, thereby deriving the Kriging equation system as
\begin{equation}
\label{eq: kriging_equation}
\begin{bmatrix}
 \bm{\Omega}_0 &  &  & & -\mathbf{I}\\
  & \bm{\Omega}_1 &  & & -\mathbf{I}\\
  &  & \ddots  & & \vdots \\
  &  &  & \bm{\Omega}_8 & -\mathbf{I}\\
 \mathbf{I}  & \mathbf{I} & \dots & \mathbf{I} & \mathbf{0}
\end{bmatrix}
\begin{bmatrix}
 \Lambda_0\\
 \Lambda_1\\
 \vdots \\
 \Lambda_8\\
 \bm{\Phi}
\end{bmatrix} = \begin{bmatrix}
\mathbf{R}_{0 \mathbf{v}} \\
\mathbf{R}_{1 \mathbf{v}} \\
\vdots \\
\mathbf{R}_{8 \mathbf{v}} \\
\mathbf{I} 
\end{bmatrix}.
\end{equation}
The optimal Kriging weight set $\{\bm{\lambda}^{*}_i\}^{8}_{i=0}$ is obtained by solving Eq. \ref{eq: kriging_equation}, one thereby can derive any elevation and normal vector by Eq. \ref{eq: kriging_pre}. Fig. \ref{fig: gnf} shows the elevation and surface normal from $\mathcal{M}$ and $\mathcal{F}_{GN}$, respectively. $\mathcal{F}_{GN}$ can offer a consecutive and smooth representation of the elevation and surface normal.
\begin{figure}[t]
    \centering
    \includegraphics[width=1.0\linewidth]{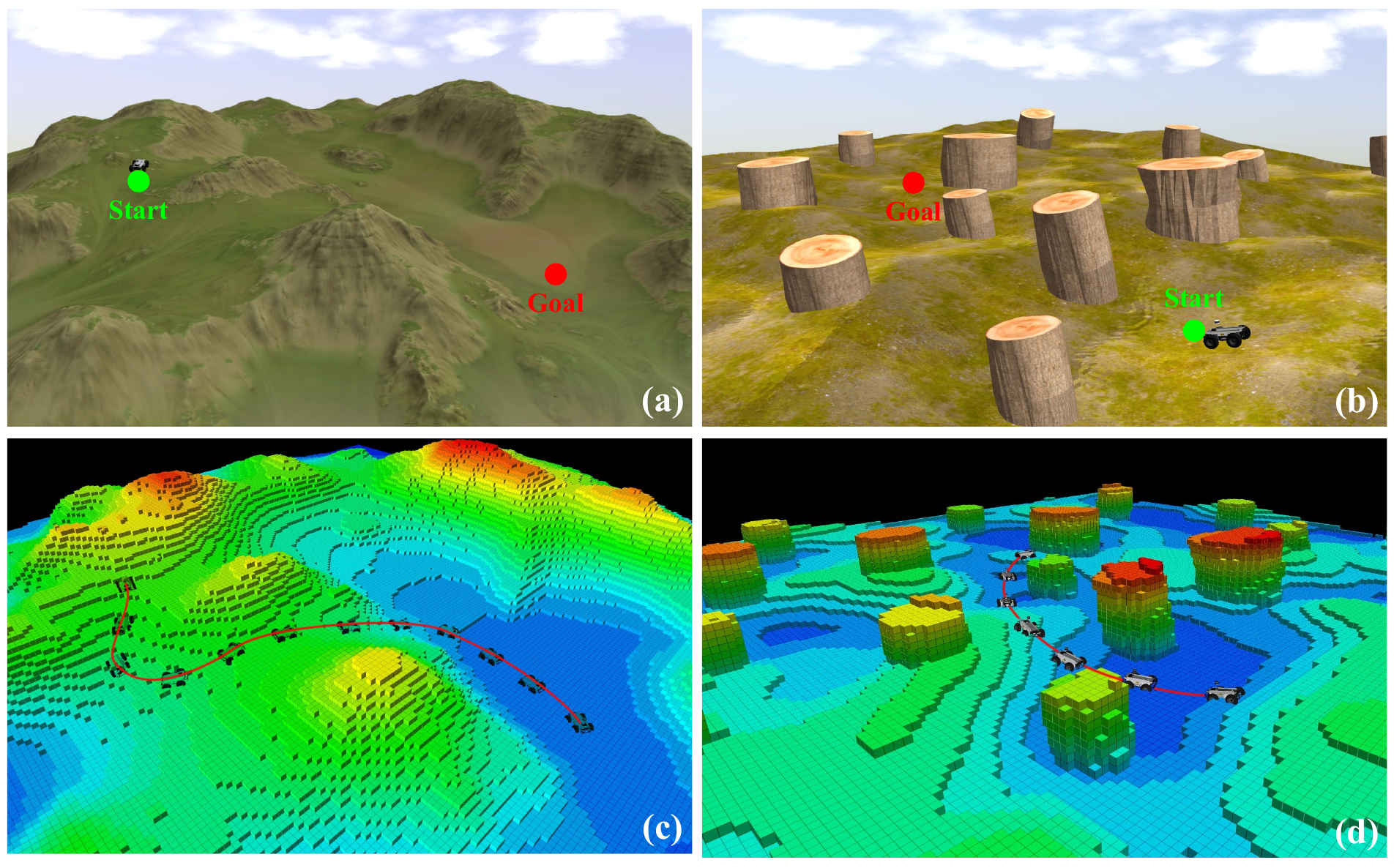}
    \caption{The navigation performance of CAP in different scenes. (a) and (b) show the topography of hilliness and forest scenes, respectively. (b) and (d) exhibit the navigation trajectories of the robot.}
    \label{fig: overall}
\end{figure}

We parameterize the trajectory $\mathcal{S}$ as the combination of the 2D state $\mathbf{s}_i$ and transit time $\bigtriangleup T_i$, which is represented by
\begin{equation}
     \mathcal{S} = \{{\mathbf{s}}_1, \Delta T_1, {\mathbf{s}}_2, \Delta T_2,...,{\mathbf{s}}_{n-1}, \Delta T_{n-1}, {\mathbf{s}}_{n}\}.
\end{equation}
$\bigtriangleup T_i$ represent the transit time from $\mathbf{s}_i$ to $\mathbf{s}_{i+1}$. One can calculate the velocity $\bm{\nu}_i$ and acceleration $\bm{a}$ of the $\mathbf{s}_i$ by
\begin{equation}
    \bm{\nu}_i = \frac{\mathbf{s}_{i+1} - \mathbf{s}_i}{\triangle T_i}, \bm{a}_i = \frac{2(\bm{\nu}_{i+1} - \bm{\nu}_i)}{\triangle T_{i+1} + \triangle T_{i}}.
\end{equation}
Inspired by Time-Elastic-Band (TEB) optimization \cite{rosmann2017teb}, we establish a trajectory optimization process as following 
\begin{equation}
\label{eq: teb_opt}
\begin{aligned}
    \mathcal{S} &= \underset{\{ \mathbf{s}_i, \triangle T_i  \}^{n}_{0}}{\mathop{\arg\min}}  \sum^{n}_{i=0} \triangle T^{2}_i, \\
    s.t.,\ &i = \{0, 1, 2, \dots, n\}, \triangle T_i \ge 0,\\
    &\begin{aligned}
        & \kappa(\mathbf{s}_i, \mathbf{s}_{i+1}) = \mathbf{0}, &&\theta \in \Theta(\mathcal{F}_{GN}, \mathbf{s}_i), \\
        & \mathbf{s}_b = \mathbf{s}_0, &&\mathbf{s}_e = \mathbf{s}_n,\\
        & \mathbf{0} \le \bm{\nu}_i \le \bm{\nu}_{m}, && \mathbf{0} \le \mathbf{a}_i \le \mathbf{a}_{m}.
    \end{aligned}
\end{aligned}
\end{equation}
Here $\mathbf{s}_b$ and $\mathbf{s}_e$ are the start and end state of the $\mathcal{S}$. $\kappa(\mathbf{s}_i, \mathbf{s}_{i+1}) = \mathbf{0}$ represents the kinematic constraints of the robot. $\Theta(\star)$ is the constraint of the traversable orientation, which is indicated by Eq. \ref{eq: traversable_orientation}. To solve Eq. \ref{eq: teb_opt} efficiently and robustly, we aim to obtain the approximate solution of Eq. \ref{eq: teb_opt} based on the graph optimization methods. Thus, Eq. \ref{eq: teb_opt} can be transformed to an unconstrained nonlinear least-squares optimization, which is written as 
\begin{equation}
\label{eq: graph_opt}
\underset{\mathcal{S}}{\mathop{\arg\min}} \sum^{n}_{i=0} \left( \gamma_T \triangle T^{2}_i + \gamma_{\Theta}f_{\Theta} +  \gamma_{\kappa}f_{\kappa} + \gamma_{\mathbf{u}}f_{\mathbf{u}} \right),
\end{equation}
where $\gamma_{\star}$ is the Karush–Kuhn–Tucker multiplier to describe the weight of penalty terms. $f_{\star}$ represents the penalty function corresponding to the constraint $\star$ in Eq. \ref{eq: teb_opt}, which is detailed in \cite{rosmann2017integrated}. Specifically, we formulate a capsizing-safety constraint based on Eq. \ref{eq: traversable_orientation}. $f_{\Theta}(\theta_i)$ is the penalty function of the capsizing-safety constraint, which can be written as
\begin{equation}
\label{eq: cap_constraint}
 f_{\Theta}(\theta_i) = \left\{\begin{matrix}
\begin{aligned}
& \infty , &&L \ge \frac{\sqrt[]{w_2+l_2} }{2}   \\
& \| \theta_i - \theta_{c} \|^{2}, &&\text{others} 
\end{aligned}
\end{matrix}\right. \end{equation}
$\theta_{c}$ points to the range center of the two-sided inequality constraint governing $\theta$ in Eq. \ref{eq: traversable_orientation}. Eq. \ref{eq: cap_constraint} indicates that the robot is expected to orient to the center of traversable orientation to ensure tip-over stability. We use the Levenberg-Marquardt (LM) algorithm to solve the optimization function of Eq. \ref{eq: graph_opt} and finally obtain the trajectory $\mathcal{S}$ accounting for multiple constraints.

\begin{figure}[t]
    \centering
    \includegraphics[width=1.0\linewidth]{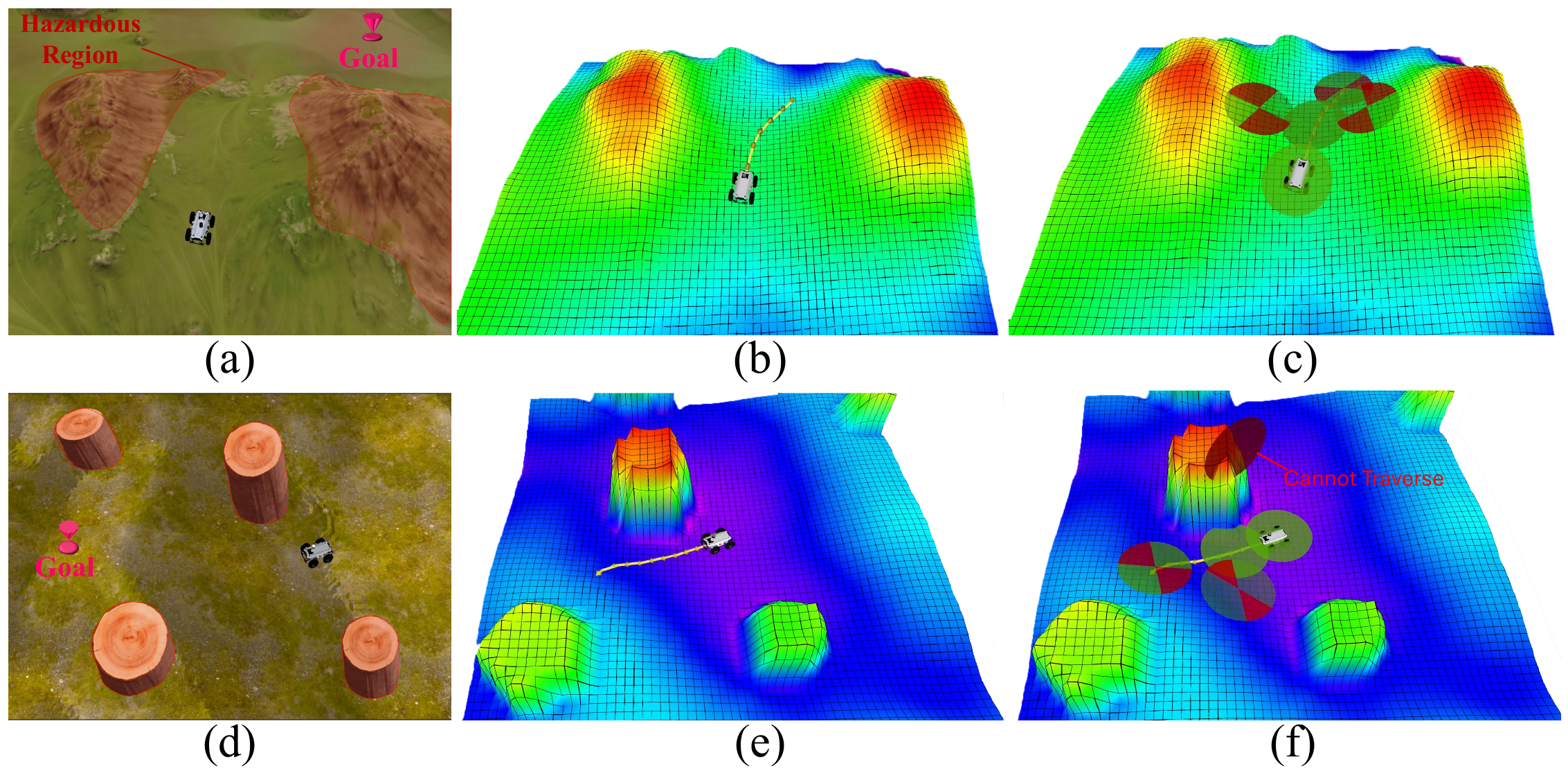}
    \caption{Snapshots of the navigation in the hilliness and forest scenes. (a) and (d) are the environmental snapshots during navigation. (b) and (e) show the trajectories generated by CAP while (c) and (f) exhibit the traversable orientation by colored pie chart. }
    \label{fig: exp_traversable}
\end{figure}

\section{Experiments and Evaluation}
\label{sec: experiments}
\subsection{Overall Performance}
To demonstrate the efficacy and effectiveness of our method in environments with uneven terrain, we conduct experiments in simulation environments. All simulation experiments are employed in an ROG laptop with an Intel Core i7-10875H CPU and 16 GB memory. The simulated robot adopts differential-drive chassis and the size is $[w, l, h]=[0.7, 0.93, 0.35](\text{m})$. We exploit Robot Operating System (ROS) Noetic software and Gazebo simulator to implement our method and build simulation scenes, respectively.

The simulation scenes and navigation configurations are as shown in Fig. \ref{fig: overall}(a)-(b). The trajectory results are shown in Fig. \ref{fig: overall}(c)-(d). The trajectories primarily follow flat and safe terrain, while climbing hills conditioned on maintaining self-stability of the robot. These trajectories demonstrate that the proposed method enables the robot to prioritize locomotion on flat terrains while avoiding hazardous topographies, such as unstable inclines and non-traversable obstacles. 
\begin{table}[b]
\centering
\caption{Evaluation statistics of navigation performance with different planners}
\label{tab: comparison}
\renewcommand\arraystretch{1.2} 
\tabcolsep=0.1cm 
\begin{tabular}{c|cc||cc||cc}
\hline
\multirow{2}{*}{Method} & \multicolumn{2}{c||}{\textbf{Hybrid A*}}      & \multicolumn{2}{c||}{\textbf{PF-RRT}}         & \multicolumn{2}{c}{\textbf{Target-Only}}     \\ \cline{2-7} 
                        & $\Upsilon$ {[}rad{]} & $\mathcal{T}$ {[}s{]} & $\Upsilon$ {[}rad{]} & $\mathcal{T}$ {[}s{]} & $\Upsilon$ {[}rad{]} & $\mathcal{T}$ {[}s{]} \\ \hline
\textbf{Ours}           & \textbf{0.746}                & 17.36                 & \textbf{0.647}                & 18.69                 & \textbf{0.850}                & \textbf{19.54}                 \\
\textbf{Teb-Planner}    & 0.990                & \textbf{15.94}                 & 1.046                & \textbf{16.91}                 & \multicolumn{2}{c}{Fail}                     \\
\textbf{Uneven-Planner} & 0.909                & 17.92             & 1.062                &    17.10             & \multicolumn{2}{c}{Fail}                     \\
\textbf{PUTN}           & {0.880}                     & {19.53}                     & 1.005                & 27.38                 & \multicolumn{2}{c}{Fail}                     \\ \hline
\end{tabular}
\end{table}

To highlight the role of the proposed traversable orientation, the navigation snapshots are captured in the hilliness and forest scene in Fig. \ref{fig: overall}(a)-(b). Fig. \ref{fig: exp_traversable}(a) and (d) exhibits the surrounding topographies of the robot, while Fig. \ref{fig: exp_traversable}(b) and (e) showcase the generated trajectories of our method. One can see that the trajectories pass through the flat area and avoid the rough surface. This is owing to the trajectory optimization with capsizing-safety constraints. In Fig. \ref{fig: exp_traversable}(c) and (f), we exhibit the traversability of orientations at the trajectory and its two-side regions. The green and red pie charts represent the traversable and non-traversable orientations, respectively. The results demonstrate that the CAP can find the most feasible trajectory in the surrounding regions, while the trajectory is satisfied with the capsizing-safety constraints.



\begin{figure}[t]
    \centering
    \includegraphics[width=1\linewidth]{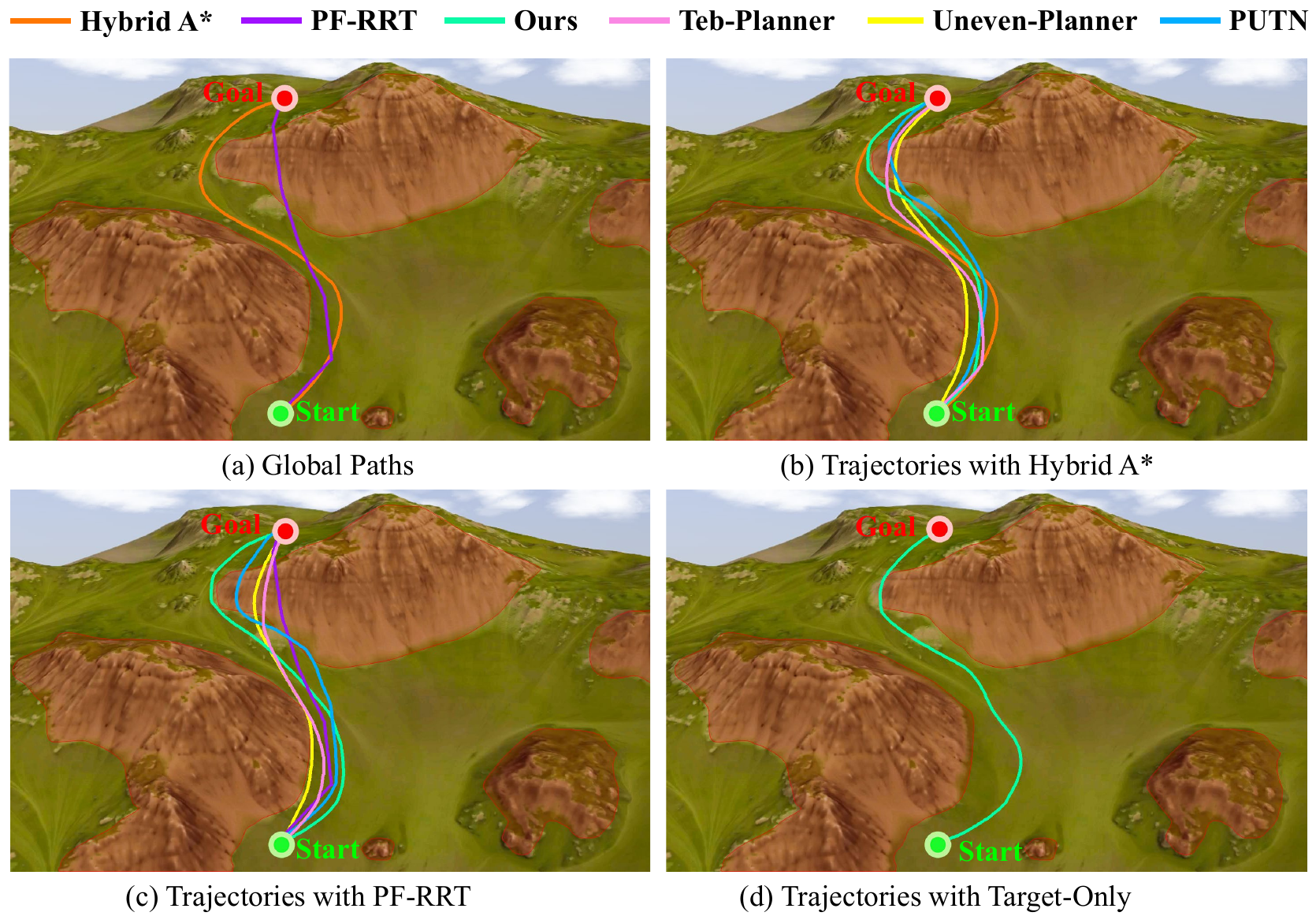}
    \caption{Trajectory Comparison under diferent global paths. (d) shows our trajectory obtained with a target-only condition, where other methods fail to navigate due to tip-over. The red regions indicate hazardous regions.}
    \label{fig: compare}
\end{figure}

\subsection{Comparison with Benchmark}
In this section, we evaluate the navigation performance of our method by comparison with \textbf{Teb-Planner} \cite{rosmann2017teb}, \textbf{PUTN} \cite{jian2022putn} and \textbf{Uneven-Planner} \cite{xu2023uneven_planner}. The maximum velocity and acceleration are set to $v_{max} = 0.8m/s$ and $a_{max} = 0.5m/s^2$, respectively. 
\begin{figure}[t]
    \centering
    \includegraphics[width=0.9\linewidth]{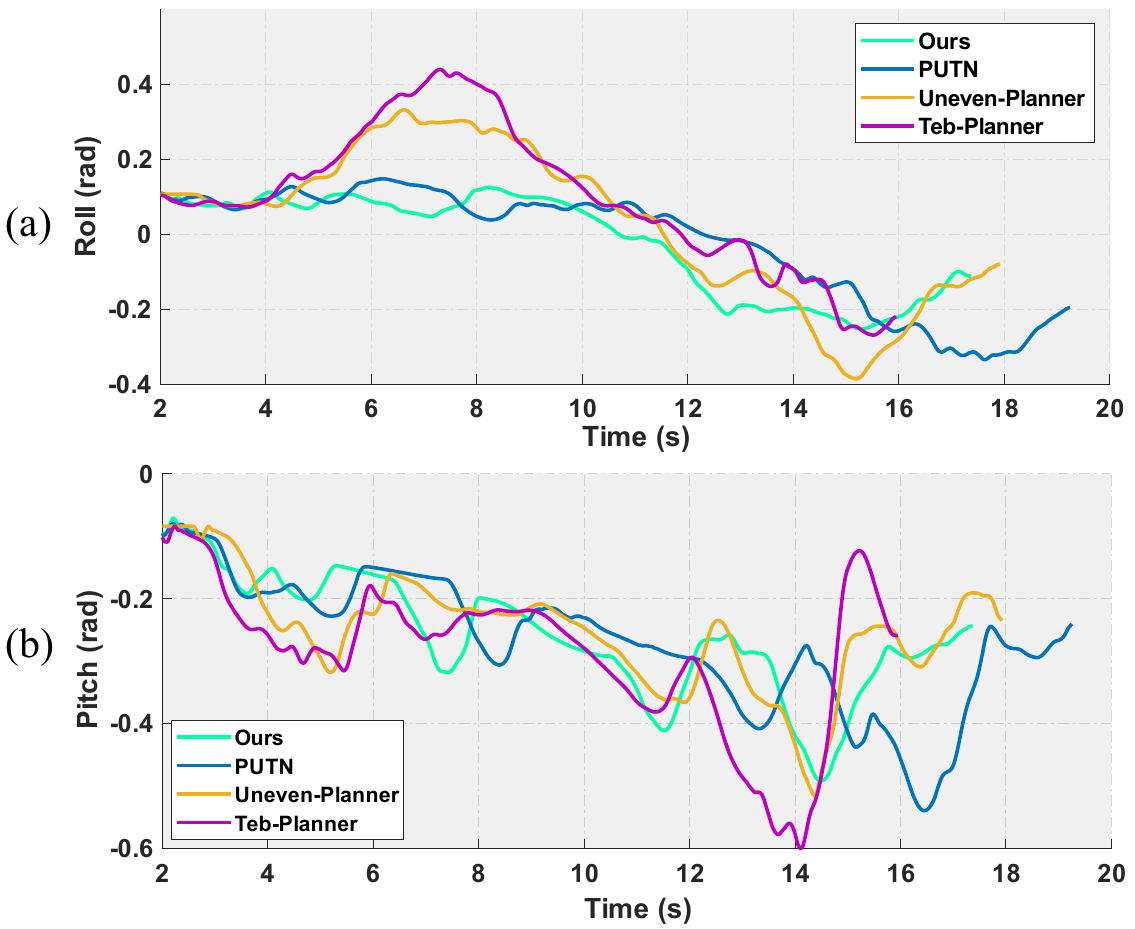}
    \caption{Variation curves of the roll and pitch angles during navigation.}
    \label{fig: comp_angle}
\end{figure}

The navigation trajectories are firstly compared in the hilliness scene shown in Fig. \ref{fig: compare}(a). We leverage different global paths to guide the robot to reach the goal on the hill's top, these global paths are obtained by Hybrid A* \cite{xu2023uneven_planner}, PF-RRT \cite{jian2022putn}, which is exhibited in Fig. \ref{fig: compare}(a). Target-Only means that we only offer a target to the planner without a global path. The comparison results are shown in Fig. \ref{fig: compare}(b)-(d). Our method successfully performs trajectory planning for both different global paths and target-only scenarios. In addition, our trajectories avoid the hazardous regions, which indicates the high performance in rejecting tip-over of our method. However, other methods fail to plan trajectories with the different global paths. Given target-only, other methods cannot generate feasible trajectories without guidance from the global paths. The results demonstrate that CAP is robust to variations in the global path and can choose to navigate to a safe surface.
\begin{figure}[b]
    \centering
    \includegraphics[width=1.0\linewidth]{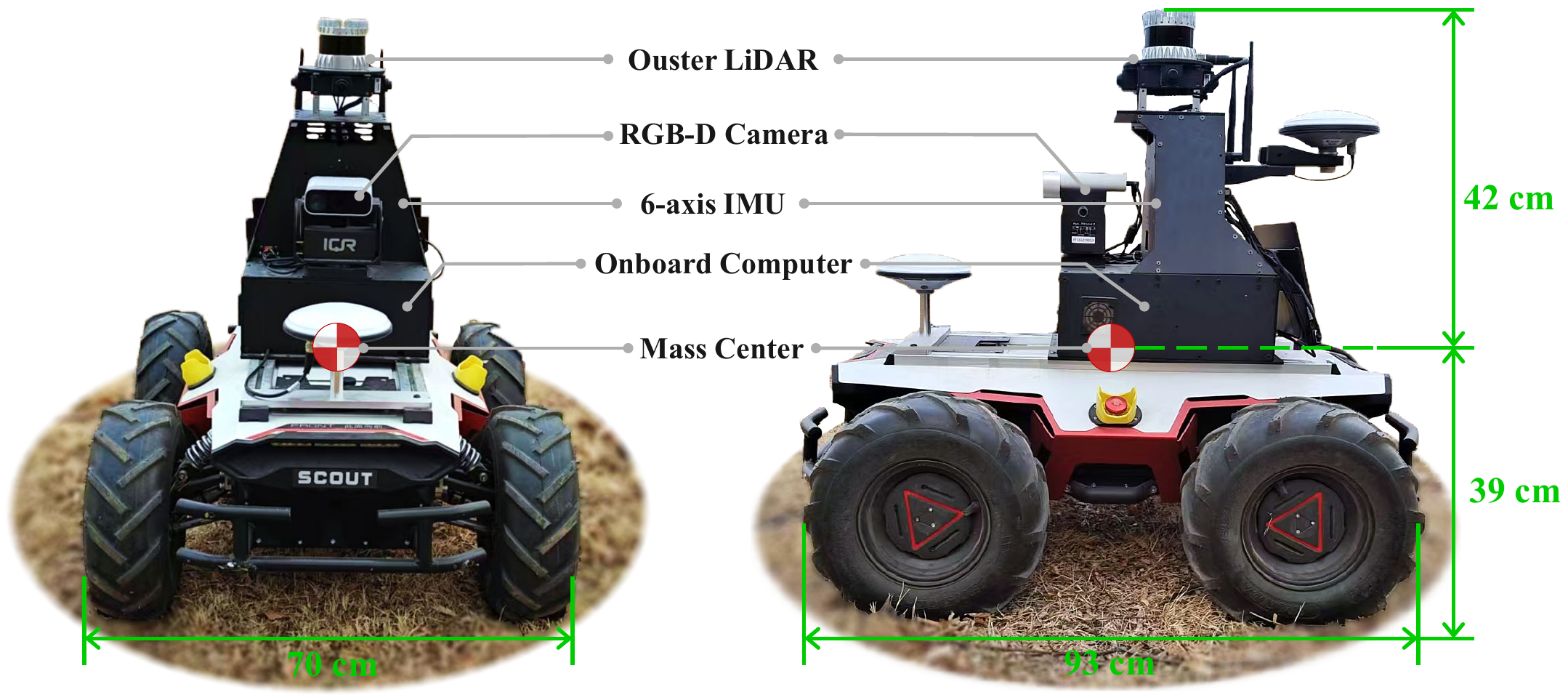}
    \caption{Hardware platform of the Scout 2.0 UGV.}
    \label{fig: hardware}
\end{figure}



Subsequently, we compare the stability and efficiency between our method and the benchmarks. Tab. \ref{tab: comparison} shows the statistics of navigation time and pose variation. $\mathcal{T}$ and $\Upsilon$ are the navigation time and the sum of the peak absolute deviations in pitch and roll angles, which are exploited to assess the efficiency and stability of navigation, respectively. The statistics results show that \textbf{Ours} method has the minimal $\Upsilon$, indicating the high performance in maintaining the self-stability. \textbf{Teb-Planner} achieves the shortest $\mathcal{T}$ with guidance of global plans, but it experiences large variations in $\Upsilon$ since it lacks consideration of the safety constraint on uneven terrain. Fig. \ref{fig: comp_angle} showcases the curve of the roll and pitch angle of the robot during the navigation in Fig. \ref{fig: compare}(b). The results are the mean values of experiments by 5 times. \textbf{Ours} exhibits the least fluctuation across the entire trajectory in Fig. \ref{fig: comp_angle}(a)-(b), which illustrates that our method is capable of generating safer and more stable trajectories with the assistance of the proposed capsizing-safety constraint.

However, both \textbf{PUTN} and \textbf{Uneven-Planner} exhibit large variations in roll and pitch angles compared to \textbf{Ours}, indicating that the robot suffers from larger bumps and becomes unstable when navigating on the ground. This instability may compromise the safety and reliability of the robot. The suboptimal performance of \textbf{PUTN} can be attributed to the absence of the transit time constraint in the trajectory optimization and unsafe sampling path. It leads to increased navigation times in complex environments, particularly in target-only cases.
In addition, the performance of \textbf{Uneven-Planner} highly relies on the quality of the global path. When the global path is suboptimal, the optimizer is easy to become trapped in local optima and the planner generates an unfeasible trajectory. Furthermore, the absence of a replanning mechanism and consideration for the tip-over stability degrades its stability and safety in challenging environments.

\begin{figure*}[t]
    \centering
    \includegraphics[width=1.0\textwidth]{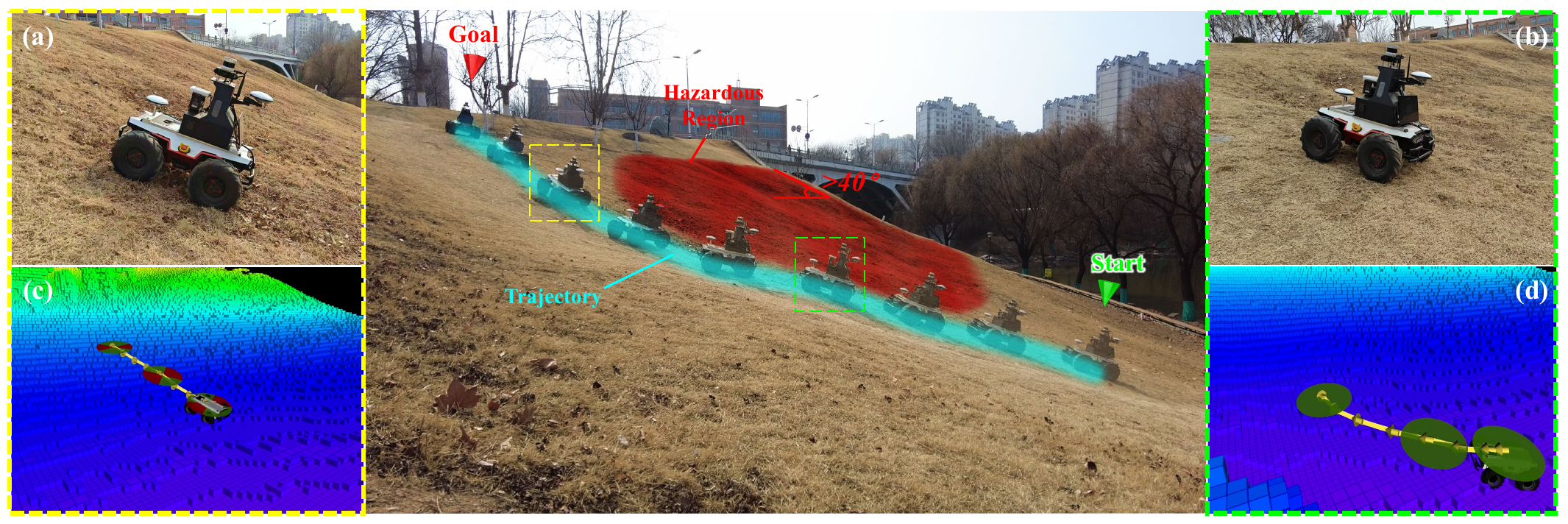}
    \caption{Autonomous navigation with CAP in the real world. The central figure shows the navigation process in real-world conditions. The red region indicates the hazardously steep slope that cannot be traversed by the robot. (a) and (b) showcase snapshots of the navigation at locations indicated by yellow and green squares, respectively. (c) and (d) exhibit the generated trajectories and corresponding traversable orientation for the snapshots.}
    \label{fig: real_world}
\end{figure*}

\subsection{Real-World Experiments}
To verify the navigation performance of our method in the real world, we implement CAP on a Scout 2.0 UGV, whose hardware configuration is shown in Fig. \ref{fig: hardware}. The robot system consists of an Ouster OS-32 LiDAR, a Microsoft Kinect RGB-D camera, 6-axis IMU, etc. The onboard computer of the robot is equipped with a NVIDIA Jetson Orin NX module and 32 GB RAM. 

The real-world environment is exhibited in Fig. \ref{fig: real_world}. The localization result is from a LiDAR SLAM algorithm, ROLO-SLAM \cite{wang2025rolo}. The central figure of Fig. \ref{fig: real_world} shows the navigation configuration and trajectory. The red regions are the steep slopes that the robot cannot traverse. The start and goal points are located at the bottom and top of the steep slope, respectively. Notably, Our method operates under the absence of a global path. As shown in the trajectory result, the robot chooses to move on the gentle terrain beside the red regions, which demonstrates that the proposed planner can generate a feasible trajectory while avoiding rough terrain. 

Fig. \ref{fig: real_world}(a)-(b) exhibit the snapshots of the navigation at the location marked by the blue and orange square, respectively. The trajectory of our method is shown in Fig. \ref{fig: real_world}(c)-(d), which is generated on relatively gentle terrain. This highlights the ability of our method to generate safe and feasible trajectories on rough terrain. This result is attributed to the proposed traversable orientation. One can see that all orientations along the trajectory lie within the range of traversable orientation, which indicates that the trajectory satisfies the proposed capsizing-safety constraint and the constraint effectively guides the generation of feasible trajectories in uneven environments.

\section{Conclusion and Future Work}
\label{sec: conclusion}
In this paper, we presented a novel trajectory planner, CAP, for the wheeled robots navigating uneven terrain, focusing on preventing tip-overs and ensuring stability. By the stability pyramid model, we derive a traversable orientation under varying terrain conditions, which is then converted into a capsizing-safety constraint and integrated into the trajectory optimization process. A robust graph-based solver is then formulated to optimize the trajectory for balancing multiple constraints to obtain smooth and safe trajectories. Through extensive simulations and real-world experiments, we demonstrated that our method outperforms existing state-of-the-art approaches, offering higher safety and stability while navigating rough terrain.

In future work, we will focus on improving terrain surface perception and optimizing trajectory smoothness, particularly in off-road environments, to enhance the robustness and efficiency of autonomous robots in rough terrains.

\addtolength{\textheight}{-12cm}   






\bibliographystyle{ieeetr}
\bibliography{ref}

\begin{thebibliography}{10}

\bibitem{wang2025rolo}
Y.~Wang, B.~Ren, X.~Zhang, P.~Wang, C.~Wang, R.~Song, Y.~Li, and M.~Q.-H. Meng.
\newblock {ROLO-SLAM: rotation-optimized LiDAR-only SLAM in uneven terrain with ground vehicle}.
\newblock \emph{Journal of Field Robotics}, 42(3):880--902, 2025.

\bibitem{xiao2021learning}
X.~Xiao, J.~Biswas, and P.~Stone.
\newblock {Learning inverse kinodynamics for accurate high-speed off-road navigation on unstructured terrain}.
\newblock \emph{IEEE Robotics and Automation Letters}, 6(3):6054--6060, 2021.

\bibitem{siva2024self}
S.~Siva, M.~Wigness, J.~G. Rogers, L.~Quang, and H.~Zhang.
\newblock {Self-reflective terrain-aware robot adaptation for consistent off-road ground navigation}.
\newblock \emph{The International Journal of Robotics Research}, 43(7):1003--1023, 2024.

\bibitem{yu2024real}
S.~Yu, C.~Shen, J.~Dallas, B.~I. Epureanu, P.~Jayakumar, and T.~Ersal.
\newblock {A Real-Time Terrain-Adaptive Local Trajectory Planner for High-Speed Autonomous Off-Road Navigation on Deformable Terrains}.
\newblock \emph{IEEE Transactions on Intelligent Transportation Systems}, 2024.

\bibitem{shen2023efficient}
C.~Shen, S.~Yu, B.~I. Epureanu, and T.~Ersal.
\newblock {An efficient global trajectory planner for highly dynamical nonholonomic autonomous vehicles on 3-D terrains}.
\newblock \emph{IEEE Transactions on Robotics}, 40:1309--1326, 2023.

\bibitem{wang2023towards}
J.~Wang, L.~Xu, H.~Fu, Z.~Meng, C.~Xu, Y.~Cao, X.~Lyu, and F.~Gao.
\newblock {Towards efficient trajectory generation for ground robots beyond 2d environment}.
\newblock In \emph{2023 IEEE International Conference on Robotics and Automation (ICRA)}, pages 7858--7864, 2023.

\bibitem{cai2023probabilistic}
X.~Cai, M.~Everett, L.~Sharma, P.~R. Osteen, and J.~P. How.
\newblock {Probabilistic traversability model for risk-aware motion planning in off-road environments}.
\newblock In \emph{2023 IEEE/RSJ International Conference on Intelligent Robots and Systems (IROS)}, pages 11297--11304, 2023.

\bibitem{atas2022elevation}
F.~Atas, G.~Cielniak, and L.~Grimstad.
\newblock {Elevation state-space: Surfel-based navigation in uneven environments for mobile robots}.
\newblock In \emph{2022 IEEE/RSJ International Conference on Intelligent Robots and Systems (IROS)}, pages 5715--5721, 2022.

\bibitem{rosmann2017integrated}
C.~R{\"o}smann, F.~Hoffmann, and T.~Bertram.
\newblock {Integrated online trajectory planning and optimization in distinctive topologies}.
\newblock \emph{Robotics and Autonomous Systems}, 88:142--153, 2017.

\bibitem{rosmann2017teb}
C.~R{\"o}smann, F.~Hoffmann, and T.~Bertram.
\newblock {Kinodynamic trajectory optimization and control for car-like robots}.
\newblock In \emph{2017 IEEE/RSJ International Conference on Intelligent Robots and Systems (IROS)}, pages 5681--5686, 2017.

\bibitem{oliver1990kriging}
Anonymous.
\newblock {Kriging: a method of interpolation for geographical information systems}.
\newblock \emph{International Journal of Geographical Information System}, 4(3):313--332, 1990.

\bibitem{tufts1982singular}
D.~Tufts and R.~Kumaresan.
\newblock {Singular value decomposition and improved frequency estimation using linear prediction}.
\newblock \emph{IEEE Transactions on Acoustics, Speech, and Signal Processing}, 30(4):671--675, 1982.

\bibitem{hariprasath2024path}
V.~Y. Hariprasath, S.~Mukesh, T.~N. Aruna, R.~Gayana, J.~Akshya, M.~D. Choudhry, and M.~Sundarrajan.
\newblock {Path Planning Solution for Intelligent Robots Using Ex* RRT Algorithm in Disaster Relief}.
\newblock In \emph{2024 5th International Conference on Recent Trends in Computer Science and Technology (ICRTCST)}, pages 413--418, 2024.

\bibitem{chang2022lamp}
Y.~Chang, K.~Ebadi, C.~E. Denniston, M.~F. Ginting, A.~Rosinol, A.~Reinke, M.~Palieri, J.~Shi, A.~Chatterjee, B.~Morrell, et~al.
\newblock {LAMP 2.0: A robust multi-robot SLAM system for operation in challenging large-scale underground environments}.
\newblock \emph{IEEE Robotics and Automation Letters}, 7(4):9175--9182, 2022.

\bibitem{wang2024hap}
Y.~Wang, N.~Du, Y.~Qin, X.~Zhang, R.~Song, and C.~Wang.
\newblock {History-Aware Planning for Risk-free Autonomous Navigation on Unknown Uneven Terrain}.
\newblock In \emph{2024 IEEE International Conference on Robotics and Automation (ICRA)}, pages 7583--7589, 2024.

\bibitem{xu2023uneven_planner}
L.~Xu, K.~Chai, Z.~Han, H.~Liu, C.~Xu, Y.~Cao, and F.~Gao.
\newblock {An Efficient Trajectory Planner for Car-Like Robots on Uneven Terrain}.
\newblock In \emph{2023 IEEE/RSJ International Conference on Intelligent Robots and Systems (IROS)}, pages 2853--2860, 2023.

\bibitem{jian2022putn}
Z.~Jian, Z.~Lu, X.~Zhou, B.~Lan, A.~Xiao, X.~Wang, and B.~Liang.
\newblock {Putn: A plane-fitting based uneven terrain navigation framework}.
\newblock In \emph{2022 IEEE/RSJ International Conference on Intelligent Robots and Systems (IROS)}, pages 7160--7166, 2022.

\bibitem{abe2024gpr}
A.~Leininger, M.~Ali, H.~Jardali, and L.~Liu.
\newblock {Gaussian Process-based Traversability Analysis for Terrain Mapless Navigation}.
\newblock In \emph{2024 IEEE International Conference on Robotics and Automation (ICRA)}, pages 10925--10931, 2024.

\bibitem{wang2020stable}
C.~Wang, M.~Xia, and M.~Q.-H. Meng.
\newblock {Stable autonomous robotic wheelchair navigation in the environment with slope way}.
\newblock \emph{IEEE Transactions on Vehicular Technology}, 69(10):10759--10771, 2020.

\bibitem{kunhe2005lmpc}
F.~K{\"u}nhe, J.~Gomes, and W.~Fetter.
\newblock {Mobile robot trajectory tracking using model predictive control}.
\newblock In \emph{II IEEE latin-american robotics symposium}, volume~51, page~5, 2005.

\bibitem{fankhauser2018elevation}
P.~Fankhauser, M.~Bloesch, and M.~Hutter.
\newblock {Probabilistic terrain mapping for mobile robots with uncertain localization}.
\newblock \emph{IEEE Robotics and Automation Letters}, 3(4):3019--3026, 2018.

\end{thebibliography}

\end{document}